\documentclass[10pt,twocolumn,letterpaper]{article}

\usepackage{cvpr}              

\usepackage{xcolor}         
\usepackage{graphicx}
\usepackage{epstopdf}
\usepackage{multirow}
\usepackage{colortbl}
\usepackage[pagebackref=true,breaklinks=true,letterpaper=true,colorlinks,bookmarks=false]{hyperref}
\usepackage{subcaption}
\usepackage{amsmath} 
\usepackage{amssymb}            
\usepackage{mathrsfs}            
\usepackage{bm}
\usepackage{wrapfig}
\usepackage{bbm}

\usepackage{algorithm}
\usepackage{algpseudocode}
\usepackage{marvosym}
\usepackage{inconsolata}

\makeatletter
\newcommand*\bigcdot{\mathpalette\bigcdot@{.5}}
\newcommand*\bigcdot@[2]{\mathbin{\vcenter{\hbox{\scalebox{#2}{$\m@th#1\bullet$}}}}}
\makeatother

\definecolor{cvprblue}{rgb}{0.21,0.49,0.74}
\usepackage{hyperref}
\definecolor{c2}{HTML}{FBD9BD}
\definecolor{c3}{HTML}{fe793d}
\definecolor{c4}{HTML}{eedeb0}
\definecolor{pp}{HTML}{BC7FCD}
\definecolor{bb}{HTML}{CDE8E5}
\definecolor{rouse}{rgb}{0.981,0.961,0.941}

\def\paperID{} 
\def\confName{CVPR}
\def\confYear{2026}

\title{
Nested Unfolding Network for Real-World Concealed Object Segmentation
}

\author{Chunming He$^{1,*}$\,,
        Rihan Zhang$^{1,*}$\,,
        Dingming Zhang$^{1}$\,,
        Fengyang Xiao$^{1}$\,,  \\
            {Deng-Ping Fan}$^{2}$ \,, 
        and {Sina Farsiu}$^{1,\dagger}$\\
        $^1$Duke University,
	$^2$Nankai University,  
\\
$*$ Equal Contribution, $\dagger$ Corresponding Author, Contact: chunming.he@duke.edu.
}

\begin{document}

\twocolumn[{
\maketitle
\vspace{-12mm}
\begin{center}
\includegraphics[width=\textwidth]{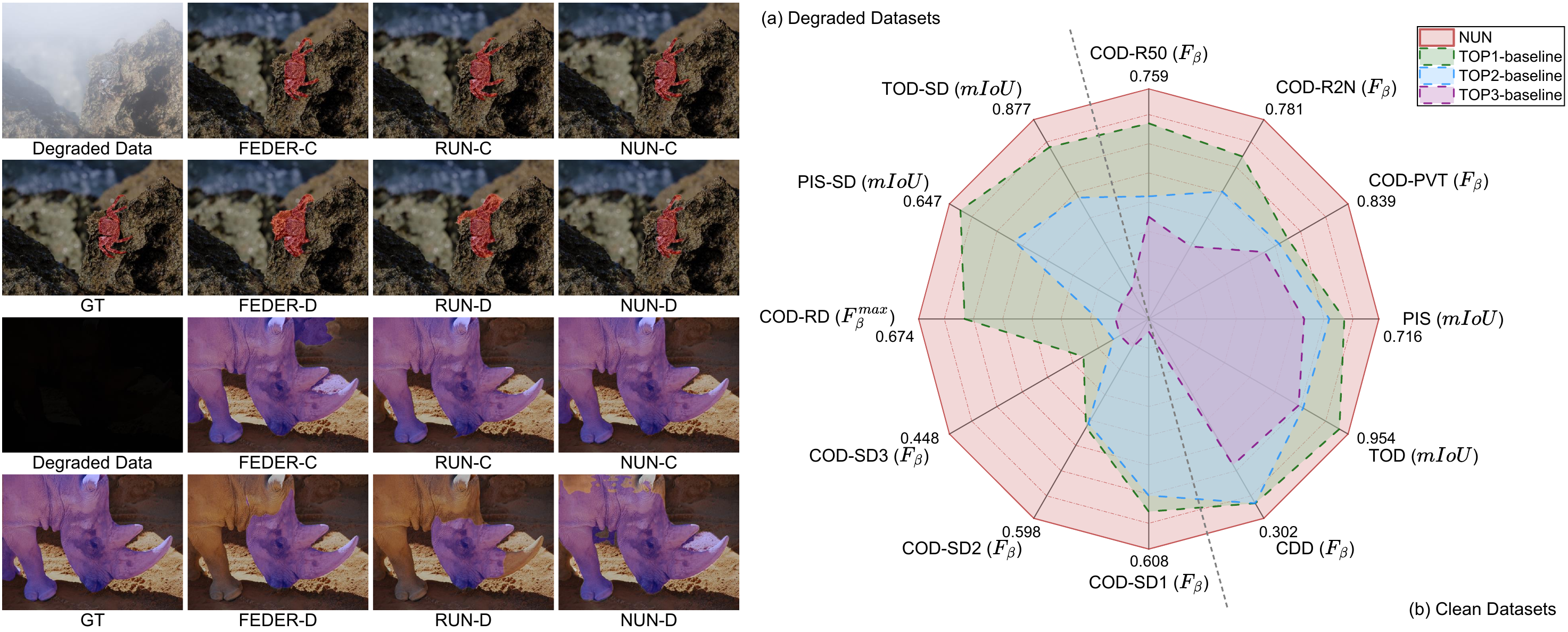}\vspace{-3mm} 
\captionof{figure}{Performance on clean and degraded COS data. Left: Concealed object masks are highlighted in {\color{red}red} and {\color{purple}purple}, overlaid on the original clean data for visual clarity. ``-C'' and ``-D'' mean segmentation on clean and degraded samples.
Right: The radar chart compares NUN with SOTAs over 12 COS tasks, where TOP1–TOP3 are composite baselines with the top metric scores per task. SD and RD denote synthetic and real-world degradation. Our NUN attains consistently superior results, especially under degradation.
} \label{Fig:ArchitectureCompare}
\end{center}
}]

\begin{abstract}
Deep unfolding networks (DUNs) have recently advanced concealed object segmentation (COS) by modeling segmentation as iterative foreground–background separation. However, existing DUN-based methods (\textit{e.g.}, RUN) inherently couple background estimation with image restoration, leading to conflicting objectives and requiring pre-defined degradation types, which are unrealistic in real-world scenarios.
To address this, we propose the nested unfolding network (NUN), a unified framework for real-world COS. NUN adopts a DUN-in-DUN design, embedding a degradation-resistant unfolding network (DeRUN) within each stage of a segmentation-oriented unfolding network (SODUN). This design decouples restoration from segmentation while allowing mutual refinement. Guided by a vision–language model (VLM), DeRUN dynamically infers degradation semantics and restores high-quality images without explicit priors, whereas SODUN performs reversible estimation to refine foreground and background.
Leveraging the multi-stage nature of unfolding, NUN employs image-quality assessment to select the best DeRUN outputs for subsequent stages, naturally introducing a self-consistency loss that enhances robustness. 
Extensive experiments show that NUN achieves a leading place on both clean and degraded benchmarks. Code will be released.

\end{abstract}
\setlength{\abovedisplayskip}{2pt}
\setlength{\belowdisplayskip}{2pt}

\section{Introduction}
\label{sec:intro}

Concealed object segmentation (COS) \cite{fan2020camouflaged,fan2021concealed,He2023Camouflaged,he2023strategic,he2023weaklysupervised,he2025segment,lu2023tf,lu2024mace,lu2024robust} aims to delineate objects that visually blend into their surroundings for intrinsic similarities, which is a challenging task requiring detecting subtle foreground–background differences.
Despite recent progress, existing COS methods remain vulnerable in real-world scenes, where degradations such as low light and haze obscure discriminative cues and cause localization errors (see \cref{Fig:ArchitectureCompare}). This degradation‑affected problem calls for unified modeling of low‑level restoration and high‑level segmentation.

\begin{figure}[t]
\setlength{\abovecaptionskip}{0cm}
	\centering
	\includegraphics[width=\linewidth]{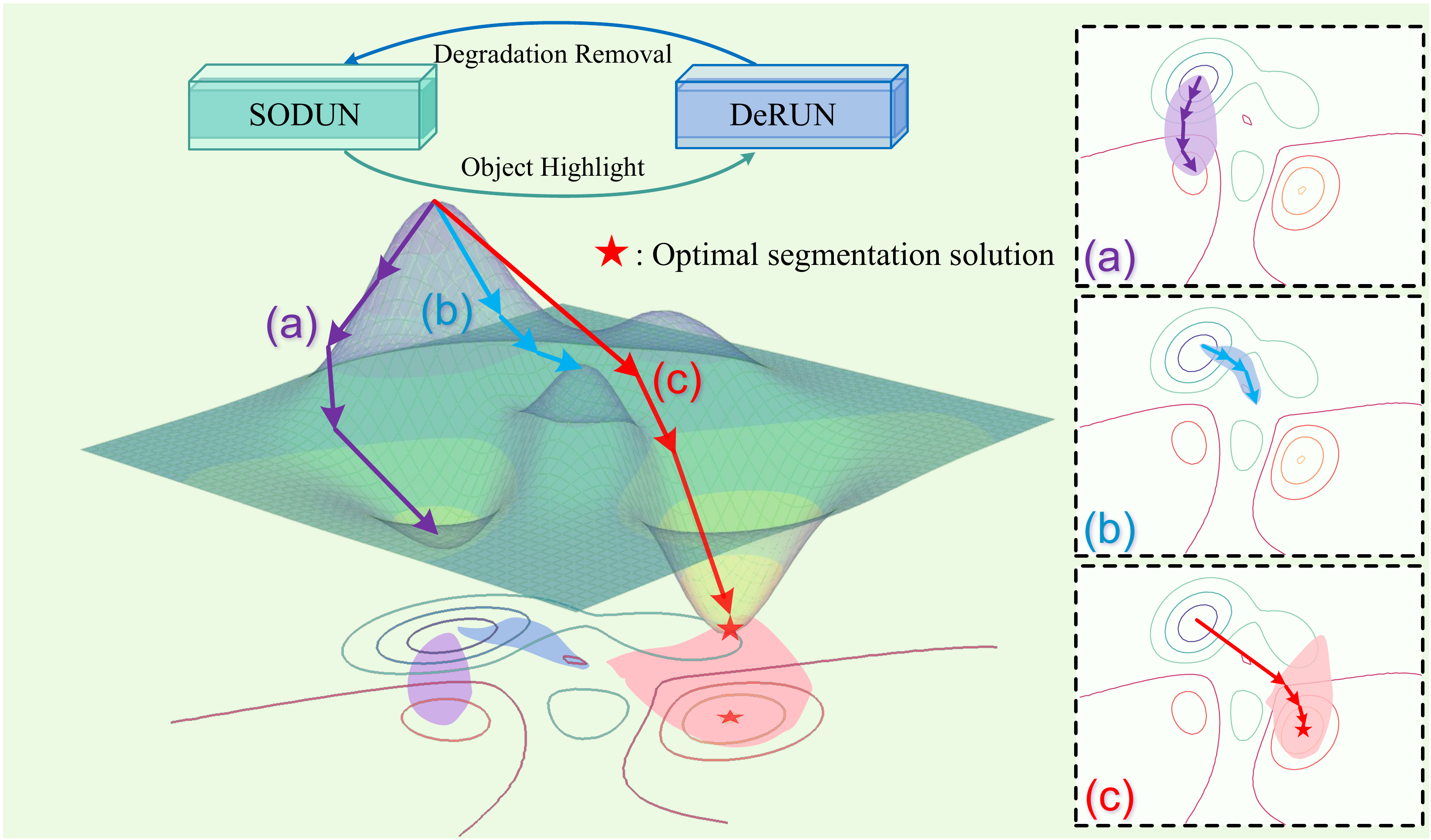}
	\caption{Conceptual motivation of the proposed NUN framework. The {\color[HTML]{00B0F0}blue} and {\color{purple}purple} lines represent restoration‑oriented and segmentation‑oriented unfolding networks, which correspondingly suffer from task inconsistency and slow convergence. In contrast, the {\color{red}red} line indicates our jointly optimized NUN, which bridges the task gap and achieves faster, more stable convergence.
    }
	\label{fig:GradientOptimization1}
	\vspace{-7mm}
\end{figure}

RUN~\cite{he2025run} introduces deep unfolding networks (DUNs) into COS, modeling segmentation as iterative foreground background separation and incorporating image restoration for enhanced robustness. However, its design inherently couples two tasks with conflicting objectives: background estimation emphasizes semantic contrast, while restoration focuses on fine-grained textures. Joint optimization leads to gradient interference and representation entanglement, impairing segmentation accuracy (as discussed in~\cref{fig:GradientOptimization1}). Moreover, the reliance on predefined degradation types limits adaptability to uncontrolled real‑world conditions.

To this end, we propose a nested unfolding network (NUN), a unified and degradation-robust framework for real-world COS. NUN adopts a DUN-in-DUN structure, in which a degradation-resistant unfolding network (DeRUN) is embedded within each stage of a segmentation-oriented unfolding network (SODUN). We employ DUN-based restoration for its interpretable iterative optimization and inherent stability under complex degradations. The multi-stage formulation enables effective interaction between restoration and segmentation, fostering mutual promotion.

In each stage, SODUN performs reversible foreground–background estimation in both mask and RGB domains, progressively refining semantic and structural details while suppressing false positives and negatives.
DeRUN, guided by a vision–language model (VLM) for degradation perception, adaptively estimates descent gradients and restores clean, degradation‑free representations without predefined degradation matrices, allowing the outer segmentation DUN to focus on robust high-level cues.

Leveraging the multi‑stage property, we introduce the bi‑directional unfolding interaction (BUI) that facilitates dynamic information exchange: segmentation outputs guide DeRUN restoration, while high-quality restorations, identified via image quality assessment, are fed back to refine subsequent segmentation stages. To ensure coherent predictions, a cross-stage consistency constraint aligns multi-stage outputs, improving stability and robustness.

Our main contributions are summarized as follows:

\noindent\textbf{(1)} We introduce NUN, a nested unfolding network for real-world COS. This is the first DUN-in-DUN structure for computer vision, enabling joint promotion of image restoration and segmentation within a unified framework. 

\noindent\textbf{(2}) We design SODUN for accurate segmentation in a reversible perspective and design a degradation-independent DeRUN guided by VLM for robust restoration under complex and unknown degradation.

\noindent\textbf{(3)} We propose a BUI mechanism that enables mutual guidance between SODUN and DeRUN. Through dynamic feature exchange and cross‑stage consistency, BUI enhances interaction and robustness, thus facilitating segmentation.

\noindent\textbf{(4)} Experiments across multiple clean and degraded benchmarks demonstrate that NUN delivers SOTA performance with computational efficiency, validating its generalization ability and scalability to broader vision tasks.

\section{Related Works}
\noindent \textbf{Concealed object segmentation}.
Early COS methods, such as SINet \cite{fan2020camouflaged}, ZoomNet \cite{pang2022zoom}, and FEDER \cite{He2023Camouflaged}, extend salient object detection frameworks by introducing task‑specific modules to extract subtle discriminative cues. However, they assume clean imaging conditions, resulting in severe performance degradation under real‑world factors such as haze \cite{fang2024real} or low illumination \cite{he2025unfoldir}. RUN \cite{he2025run} alleviates this by integrating restoration strategies and formulating COS as an optimization problem via deep unfolding. Yet, its coupling of restoration and background estimation causes optimization conflicts and sensitivity to degradation types, underscoring the need for a degradation‑robust structural design—an issue our NUN framework addresses.

\noindent \textbf{Deep unfolding network}. DUNs unfold iterative optimization algorithms into learnable multi‑stage architectures\cite{he2025unfoldir,he2025run,he2025reversible}, achieving notable success in low‑level tasks such as dehazing\cite{fang2024real} and illumination enhancement\cite{he2025unfoldir}. Recently, their adoption in high‑level vision has gained attention. In COS, RUN\cite{he2025run} pioneers this paradigm but assumes fixed degradation conditions, limiting adaptability. In contrast, our NUN introduces a nested architecture where an inner module (DeRUN) performs degradation‑agnostic restoration guided by a vision–language model, while an outer module (SODUN) carries out segmentation. This design unifies low‑ and high‑level vision processes in a theoretically grounded and practically robust framework.

\begin{figure*}[t]
\setlength{\abovecaptionskip}{0cm}
	\centering
	\includegraphics[width=\linewidth]{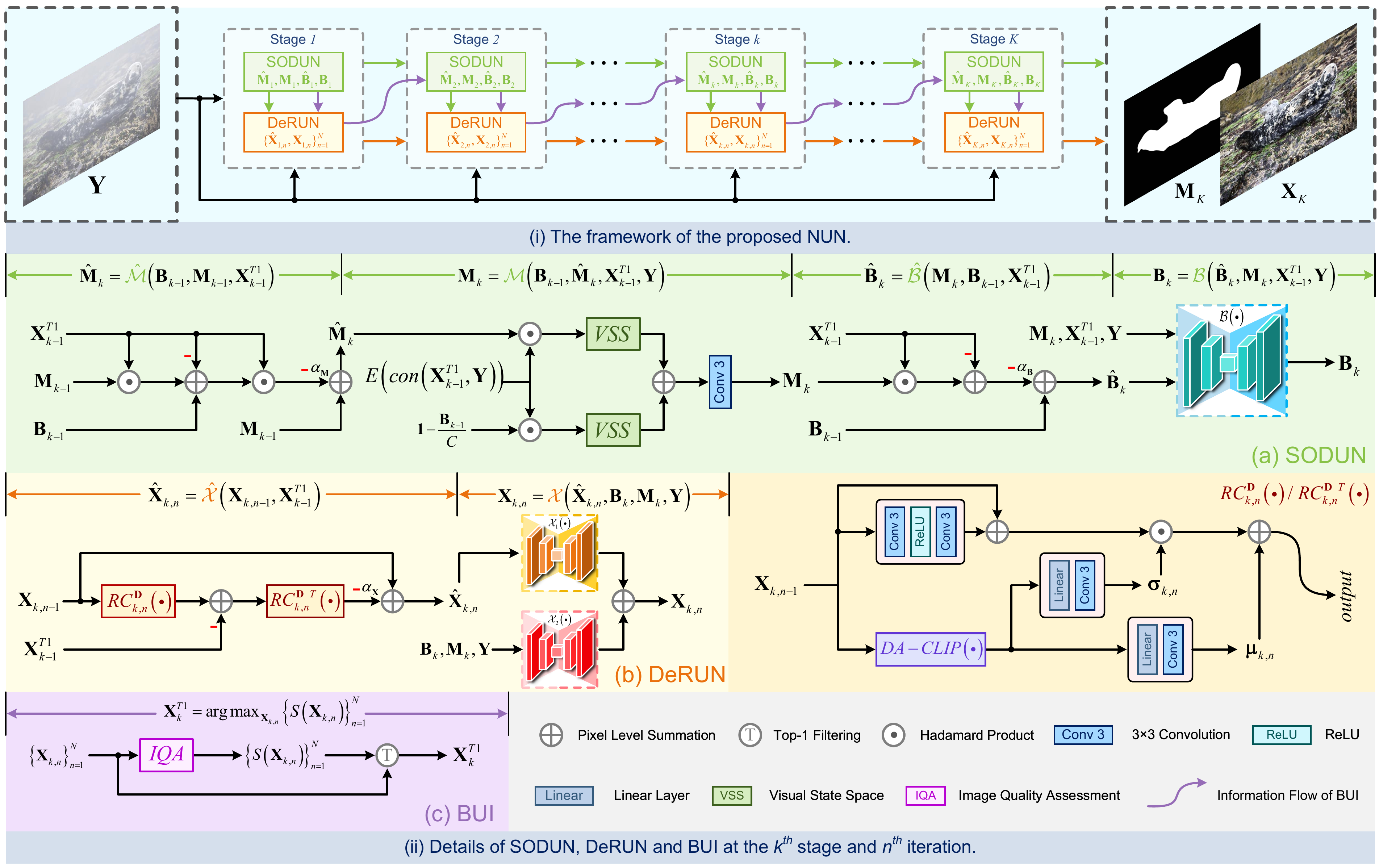}
	\caption{Framework of our proposed NUN.
    }
	\label{fig:Framework}
	\vspace{-7mm}
\end{figure*}

\section{Models and Optimization}

\subsection{Image segmentation}\label{Sec:SegModel}
\noindent\textbf{Segmentation Model}. 
A clean image $\mathbf{X}$ can be decomposed into foreground $\mathbf{F}$ and background $\mathbf{B}$ by optimizing 
\begin{equation}\label{Eq:BasicModel}
   L\left(\mathbf{F}, \mathbf{B}\right)=\frac{1}{2}\|\mathbf{X}-\mathbf{F}-\mathbf{B}\|_2^2 + \beta\varphi(\mathbf{F}) + \lambda \phi(\mathbf{B}),
\end{equation}
where $\|\bigcdot\|_2$ is an $\ell_2$-norm. $\varphi(\bigcdot)$ and $\phi(\bigcdot)$ are regularization terms with hyper-parameters $\beta$ and $\lambda$.
For segmentation, the foreground is represented as $\mathbf{F}= \mathbf{X}\odot\mathbf{M}$, with $\mathbf{M}$ and $\odot$ denoting the binary mask and dot product, respectively. Substituting into \cref{Eq:BasicModel} gives
\begin{equation}\label{Eq:BasicModel1}
   L\left(\mathbf{M}, \mathbf{B}\right)=\frac{1}{2}\|\mathbf{X}- \mathbf{X}\odot\mathbf{M} -\mathbf{B}\|_2^2 + \beta\varphi(\mathbf{M}) + \lambda \phi (\mathbf{B}).
\end{equation}

\noindent\textbf{Model Optimization}. Following the proximal‑gradient algorithm~\cite{he2025run}, we alternately optimize $\mathbf{M}$ and $\mathbf{B}$ at stage $k$:
\begin{equation}\label{Eq:SolutionM1}
       \mathbf{M}_k = \arg \underset{\mathbf{M}}{\min } \ \frac{1}{2}\|\mathbf{X}- \mathbf{X}\odot\mathbf{M} -\mathbf{B}_{k-1}\|_2^2 + \mu\psi(\mathbf{M}), 
\end{equation} 
\begin{equation}\label{Eq:SolutionB1}
       \mathbf{B}_k=\arg \underset{\mathbf{B}}{\min } \frac{1}{2}\|\mathbf{X}- \mathbf{X}\odot\mathbf{M}_k -\mathbf{B}\|_2^2 + \lambda \phi (\mathbf{B}).
\end{equation}

We first optimize $\mathbf{M}_k$, which consists of a gradient descent term and a proximal term, formulated as:
\begin{equation}\hspace{-3mm}
\begin{aligned}
    \hat{\mathbf{M}}_k &\!=\! \mathbf{M}_{k-1}-\alpha_{\mathbf{M}}\nabla m(\mathbf{M}_{k-1}), \\
    &\!=\! \mathbf{M}_{k-1}-\alpha_{\mathbf{M}}(\mathbf{X}\odot(\mathbf{X}\odot\mathbf{M}_{k-1}+\mathbf{B}_{k-1}-\mathbf{X})),
\end{aligned}
\end{equation}
\begin{equation}
    \mathbf{M}_k = \text{prox}_{\mu,\psi}(\hat{\mathbf{M}}_k),
\end{equation}
where $m(\bigcdot)$ denotes the data fidelity term of~\cref{Eq:SolutionM1}. $\hat{\mathbf{M}}_k$ is the intermediate solution and $\alpha_{\mathbf{M}}$ is a step-size parameter. Similarly, $\mathbf{B}_k$ is optimized with a step-size parameter $\alpha_{\mathbf{B}}$:
\begin{equation}
\begin{aligned}
    \hat{\mathbf{B}}_k &= \mathbf{B}_{k-1}-\alpha_{\mathbf{B}}\nabla b(\mathbf{B}_{k-1}), \\
    &= \mathbf{B}_{k-1}-\alpha_{\mathbf{B}}(\mathbf{B}_{k-1}+\mathbf{X}\odot\mathbf{M}_k-\mathbf{X}),
\end{aligned}
\end{equation}
\begin{equation}
    \mathbf{B}_k = \text{prox}_{\lambda,\phi}(\hat{\mathbf{B}}_k),
\end{equation}
where $b(\bigcdot)$ is the data fidelity term of~\cref{Eq:SolutionB1}.
\subsection{Image restoration}\label{Sec:ResModel}\vspace{-1mm}
\noindent\textbf{Restoration Model}. 
Given the degraded observation $\mathbf{Y}$, the degradation process can be formulated as:
\begin{equation}
    \mathbf{Y} = \mathbf{D} \mathbf{X}  + \mathbf{N},
\end{equation}
where $\mathbf{D}$ denotes the unknown degradation operator and $\mathbf{N}$ is the random additive noise. The restoration objective is
\begin{equation}
 L(\mathbf{X}) =  \; \frac{1}{2}\|\mathbf{Y} - \mathbf{D} \mathbf{X}\|^2_2 + \gamma \omega(\mathbf{X}),
\end{equation}
where $\omega(\bigcdot)$ regularizes the restored image with weight $\gamma$. 
\setlength{\textfloatsep}{4pt}
\begin{algorithm}[t]
	\caption{Proposed NUN Algorithm.}
	\label{Alg:NUN}
	\textbf{Input}: degraded concealed image $\mathbf{Y}$, stage number $K$, iteration number $N$ \\
	\textbf{Output}: concealed object mask $\mathbf{M}_K$, restored image $\mathbf{X}_K^{T1}$
	\begin{algorithmic}[1]
		\State Zero initialization for $\mathbf{M}_0$, $\mathbf{B}_0$. $\mathbf{X}_0^{T1}=\mathbf{Y}$, $\mathbf{X}_{0,0}=\mathbf{Y}$ 
		\For{each stage $k\in \left[1,K\right]$} 
            \State $\hat{\mathbf{M}}_k = \hat{\mathcal{M}}(\mathbf{B}_{k-1},\mathbf{M}_{k-1},\mathbf{X}_{k-1}^{T1})$,  
            \State $\mathbf{M}_k = \mathcal{M}(\mathbf{B}_{k-1}, \hat{\mathbf{M}}_k, \mathbf{X}_{k-1}^{T1},\mathbf{Y})$, 
		\State $\hat{\mathbf{B}}_k =\hat{\mathcal{B}}\left(\mathbf{M}_k, \mathbf{B}_{k-1}, \mathbf{X}_{k-1}^{T1}\right),$  
        \State ${\mathbf{B}}_k = {\mathcal{B}}\left(\hat{\mathbf{B}}_{k}, \mathbf{M}_k, \mathbf{X}_{k-1}^{T1},\mathbf{Y} \right),$ 
        \For{each iteration $n\in \left[1,N\right]$}
        \State $\hat{\mathbf{X}}_{k,n} = \hat{\mathcal{X}}(\mathbf{X}_{k,n-1}, \mathbf{X}_{k-1}^{T1})$,
        \State ${\mathbf{X}}_{k,n} = \mathcal{X}(\hat{\mathbf{X}}_{k,n}, {\mathbf{B}}_k,\mathbf{M}_k, \mathbf{Y})$,
        \EndFor
        \State $\mathbf{X}_k^{T1} = \arg\max_{\mathbf{X}_{k,n}} \{S(\mathbf{X}_{k,n})\}_{n=1}^N$.
		\EndFor
	\end{algorithmic}
\end{algorithm}

\noindent\textbf{Model Optimization}. The optimization function of the restoration problem is formulated as:
\begin{equation}\label{eq:optim}
 \mathbf{X}_k = \arg \min_{\mathbf{X}} \; \frac{1}{2}\|\mathbf{Y} - \mathbf{D} \mathbf{X}\|^2_2 + \gamma \omega(\mathbf{X}).
\end{equation}

We use the proximal gradient algorithm to optimize the restoration problem with the step-size parameter $\alpha_{\mathbf{X}}$:
\begin{equation}\label{eq:gradient1}
\begin{aligned}
     \hat{\mathbf{X}}_k &= \mathbf{X}_{k-1}-\alpha_{\mathbf{X}}\nabla x(\mathbf{X}_{k-1}),\\
     &= \mathbf{X}_{k-1}-\alpha_{\mathbf{X}} \mathbf{D}^T( \mathbf{D}\mathbf{X}_{k-1} - \mathbf{Y}),
\end{aligned}
\end{equation}
\begin{equation}\label{eq:proxim}
 {\mathbf{X}}_k = \text{prox}_{\gamma,\omega}(\hat{\mathbf{X}}_k),
\end{equation}
where $x(\bigcdot)$ is the image fidelity term of~\cref{eq:optim}.

\section{NUN}
\noindent \textbf{Motivations}.
RUN~\cite{he2025run} introduces deep unfolding networks (DUNs) into COS by modeling segmentation as iterative foreground–background separation and embedding restoration methods to handle degraded inputs. 
However, this design couples background estimation and restoration, whose conflicting goals, semantic suppression versus texture recovery, cause feature entanglement and unstable optimization. Moreover, their reliance on predefined degradation types limits applicability to real-world scenes.

To overcome these issues, we decouple restoration and segmentation into two complementary unfolding processes and adopt DUN for both, leveraging its iterative interpretability and multi-stage stability. Hence, we propose the Nested Unfolding Network (NUN), as shown in~\cref{fig:Framework,Alg:NUN}. NUN is a unified, degradation‑robust framework that adopts a DUN‑in‑DUN architecture: a degradation‑resistant network (DeRUN) nested within a segmentation‑oriented network (SODUN). In each stage, SODUN performs reversible foreground–background estimation, while DeRUN restores clean representations guided by a vision–language model (VLM) for degradation perception. A bi‑directional unfolding interaction (BUI) further facilitates reciprocal information exchange between segmentation and restoration, forming a closed collaborative loop.

\subsection{SODUN} 
Derived from the segmentation model in~\cref{Sec:SegModel}, SODUN takes the degraded image $\mathbf{Y}$ as input. 
Two modules, $\{\hat{\mathcal{M}}(\bigcdot), \mathcal{M}(\bigcdot)\}$ and $\{\hat{\mathcal{B}}(\bigcdot), \mathcal{B}(\bigcdot)\}$, simulate the gradient descent and proximal steps for segmentation mask prediction and background estimation, respectively.

\noindent\textbf{Mask prediction}. At the $k^{th}$ stage, given $\mathbf{B}_{k-1}$ and $\mathbf{M}_{k-1}$, the intermediate mask $\mathbf{M}_{k}$ is obtained as:
\begin{equation}\hspace{-3mm}\label{eq:SODUN_Mhat}
\begin{aligned}
    \hat{\mathbf{M}}_k &\!=\! \hat{\mathcal{M}}(\mathbf{B}_{k-1},\mathbf{M}_{k-1},\mathbf{Y}), \\
    &\!=\! \mathbf{M}_{k-1}\!-\!\alpha_{\mathbf{M}}(\mathbf{Y}\odot(\mathbf{Y}\odot\mathbf{M}_{k-1}\!+\!\mathbf{B}_{k-1}\!-\!\mathbf{Y})).
\end{aligned}
\end{equation}

\cref{eq:SODUN_Mhat} strictly follows the proximal gradient derivation, while all fixed parameters are made learnable to enhance generalization. With the aid of DeRUN, the enhanced result $\mathbf{X}_{k-1}$ replaces the low-quality input $\mathbf{Y}$, yielding:
\begin{equation}\hspace{-3mm}\label{eq:maskgradient}
\begin{aligned}
    \hat{\mathbf{M}}_k &\!=\! \hat{\mathcal{M}}(\mathbf{B}_{k-1},\mathbf{M}_{k-1},\mathbf{X}_{k-1}).
\end{aligned}
\end{equation}

To further refine $\hat{\mathbf{M}}$, the proximal term adopts a visual state space (VSS) module to capture non-local dependencies,  which is essential for global reasoning based on sparse discriminative cues visible in concealed settings. Following common practice~\cite{he2025run}, we employ deep encoder features $E(\mathbf{X}_{k-1})$ extracted by ResNet50~\cite{he2016deep}. To mitigate potential restoration bias, both the restored and original images are utilized. The refined mask $\mathbf{M}_k$ is calculated as:
\begin{equation}\hspace{-3mm}\label{eq:maskproximal}
\begin{aligned}
    \mathbf{M}_k&\!=\!\mathcal{M}(\mathbf{B}_{k-1}, \hat{\mathbf{M}}_k, \mathbf{X}_{k-1},\mathbf{Y}), \\
    &\!=\! conv3(VSS(\hat{\mathbf{M}}_k \odot E(con(\mathbf{X}_{k-1},\mathbf{Y}))) \\
    &\!+\!  VSS((\mathbf{1}\!-\!\mathbf{B}_{k-1}/\mathbf{C})\odot E(con(\mathbf{X}_{k-1},\mathbf{Y})))),
\end{aligned}
\end{equation}
where $con(\bigcdot)$, $VSS(\bigcdot)$, and $conv3(\bigcdot)$ denote concatenation, visual state space, and $3\times 3$ convolution, respectively. 

\noindent\textbf{Background estimation}. Given $\mathbf{M}_{k}$ and $\mathbf{B}_{k-1}$, the background $\hat{\mathbf{B}}_k$ is corrected via the gradient descent step:
\begin{equation}\hspace{-3mm}\label{eq:backgroundgradient}
\begin{aligned}
    \hat{\mathbf{B}}_k &\!=\! \hat{\mathcal{B}}(\mathbf{M}_{k},\mathbf{B}_{k-1},\mathbf{X}_{k-1}), \\
    &\!=\! \mathbf{B}_{k-1}-\alpha_{\mathbf{B}}(\mathbf{B}_{k-1}+\mathbf{X}_{k-1}\odot\mathbf{M}_k-\mathbf{X}_{k-1}).
\end{aligned}
\end{equation}

The proximal term refines this estimate using a compact three-layer U-shaped network $\mathcal{B}(\bigcdot)$~\cite{he2023HQG}:
\begin{equation}\hspace{-3mm}\label{eq:backgroundproximal}
\begin{aligned}
    {\mathbf{B}}_k &\!=\! {\mathcal{B}}(\hat{\mathbf{B}}_k,\mathbf{M}_k,\mathbf{X}_{k-1}, \mathbf{Y}). 
\end{aligned}
\end{equation}

The degraded–restored pair $\{\mathbf{X}_{k-1}, \mathbf{Y}\}$ provides valuable degradation‑related cues that assist the proximal terms in more precise mask prediction and background correction.

\noindent \textbf{Discussion}. By reversibly estimating the foreground and background in both the mask and RGB domains, SODUN encourages complementary feedback. This drives the model to focus on ambiguous boundary regions where reciprocal supervision highlights discriminative cues, mitigating false predictions and promoting segmentation results.

\begin{table*}[tbp!]
\begin{minipage}{1\textwidth}
\setlength{\abovecaptionskip}{0cm} 
		\centering
            \caption{Results on camouflaged object detection with clean data.  
            } \label{table:CODQuanti}
            \vspace{1mm}
		\resizebox{\columnwidth}{!}{
			\setlength{\tabcolsep}{1.4mm}
			\begin{tabular}{l|c|cccc|cccc|cccc|cccc} 
				\toprule
				\multicolumn{1}{c|}{}                                        & \multicolumn{1}{c|}{}                           & \multicolumn{4}{c|}{\textit{CHAMELEON} }                                                                                                                                         & \multicolumn{4}{c|}{\textit{CAMO} }                                                                                                                                             & \multicolumn{4}{c|}{\textit{COD10K} }                                                                                                                                          & \multicolumn{4}{c}{\textit{NC4K} }                                                                                                                        \\ \cline{3-18} 
				\multicolumn{1}{l|}{\multirow{-2}{*}{Methods}} & \multicolumn{1}{c|}{\multirow{-2}{*}{Backbones}} & {\cellcolor{gray!40}$M$~$\downarrow$}                                  & {\cellcolor{gray!40}$F_\beta$~$\uparrow$}                               & {\cellcolor{gray!40}$E_\phi$~$\uparrow$}                               & \multicolumn{1}{c|}{\cellcolor{gray!40}$S_\alpha$~$\uparrow$}                                   & {\cellcolor{gray!40}$M$~$\downarrow$}                                  & {\cellcolor{gray!40}$F_\beta$~$\uparrow$}                               & {\cellcolor{gray!40}$E_\phi$~$\uparrow$}                               & \multicolumn{1}{c|}{\cellcolor{gray!40}$S_\alpha$~$\uparrow$}                                   & {\cellcolor{gray!40}$M$~$\downarrow$}                                  & {\cellcolor{gray!40}$F_\beta$~$\uparrow$}                               & {\cellcolor{gray!40}$E_\phi$~$\uparrow$}                               & \multicolumn{1}{c|}{\cellcolor{gray!40}$S_\alpha$~$\uparrow$}                                   & {\cellcolor{gray!40}$M$~$\downarrow$}                                  & {\cellcolor{gray!40}$F_\beta$~$\uparrow$}                               & {\cellcolor{gray!40}$E_\phi$~$\uparrow$}                               & \multicolumn{1}{c}{\cellcolor{gray!40}$S_\alpha$~$\uparrow$}                                   \\ \midrule 
                \multicolumn{1}{l|}{FEDER~\cite{He2023Camouflaged}} & \multicolumn{1}{c|}{ResNet50}  & { {0.028}} & { {0.850}} & 0.944 & \multicolumn{1}{c|}{0.892} & {{0.070}} & {0.775} & 0.870 & \multicolumn{1}{c|}{0.802} & 0.032 & 0.715 & 0.892 & \multicolumn{1}{c|}{0.810} & {{0.046}} & {{0.808}} & {{0.900}} & {{0.842}} \\
                \multicolumn{1}{l|}{FSEL~\cite{sun2025frequency}} & \multicolumn{1}{c|}{ResNet50} & 0.029 & 0.847 & 0.941 & {{0.893}}  & {{\textbf{0.069}}} & {{0.779}} & {0.881} & {{\textbf{0.816}}} & 0.032 & 0.722 & 0.891 & {{0.822}} & 0.045 & 0.807 & 0.901 & 0.847    \\ 
                \multicolumn{1}{l|}{RUN~\cite{he2025run}}  & \multicolumn{1}{c|}{ResNet50} & {{0.027}} & {{0.855}} & {{0.952}} & {{0.895}} & 0.070 & {{0.781}} & 0.868 & 0.806 & \textbf{0.030} & {{0.747}} & {{0.903}} & {{0.827}} & \textbf{0.042} & {{0.824}} & {{0.908}} & {{0.851}}     \\
                \rowcolor{c2!20} NUN (Ours) & ResNet50 & \textbf{0.026} & \textbf{0.859} & \textbf{0.953} & \textbf{0.899} & 0.070 & \textbf{0.782} & \textbf{0.883} & 0.809 & \textbf{0.030} & \textbf{0.759} & \textbf{0.904} & \textbf{0.829} & \textbf{0.042} & \textbf{0.825} & \textbf{0.909} & \textbf{0.852} \\
                \midrule
				\multicolumn{1}{l|}{BSA-Net~\cite{zhu2022can}}            & \multicolumn{1}{c|}{Res2Net50}                  & 0.027                                 & 0.851                                 & 0.946                                 & \multicolumn{1}{c|}{0.895}                                 & 0.079                                 & 0.768                                 & 0.851                                 & \multicolumn{1}{c|}{0.796}                                 & 0.034                                 & 0.723                                 & 0.891                                 & \multicolumn{1}{c|}{0.818}                                 & 0.048                                 & 0.805                                 & 0.897                                 & 0.841                                 \\
               \multicolumn{1}{l|}{RUN~\cite{he2025run}}  & \multicolumn{1}{c|}{Res2Net50} & {0.024} & {0.879} & {0.956} & {0.907} & {\textbf{0.066}} & {0.815} & {0.905} & {{0.843}}  & {0.028} & {0.764} & {\textbf{0.914}} & {\textbf{0.849}} & {0.041} & {0.830} & {0.917} &0.859                \\
                \rowcolor{c2!20} NUN (Ours) & Res2Net50 & \textbf{0.024} & \textbf{0.883} & \textbf{0.957} & \textbf{0.908} & \textbf{0.066} & \textbf{0.822} & \textbf{0.907} & \textbf{0.845} & \textbf{0.027} & \textbf{0.781} & \textbf{0.914} & \textbf{0.849} & \textbf{0.040} & \textbf{0.836} & \textbf{0.918} & \textbf{0.863}
                \\ 
                \midrule
                CamoDiff~\cite{sun2025conditional} & PVT V2 & 0.022 & 0.868 & 0.952 & 0.908 & \textbf{0.042} & 0.853 & 0.936 & 0.878 & \textbf{0.019} & 0.815 & 0.943 & 0.883 & \textbf{0.028} & 0.858 & 0.942 & 0.895 \\
                \multicolumn{1}{l|}{RUN~\cite{he2025run}} &  \multicolumn{1}{c|}{PVT V2} & {\textbf{0.021}} & {0.877} & {\textbf{0.958}} & {\textbf{0.916}} & 0.045 &  {0.861} &  {0.934} & {0.877} & {0.021} &  0.810 & {0.941} &  {0.878} & {0.030} & {0.868} & {0.940} & {0.892} \\ 
                 \rowcolor{c2!20} NUN (Ours) & PVT V2 & \textbf{0.021} & \textbf{0.888} & \textbf{0.958} & \textbf{0.916} & \textbf{0.042} & \textbf{0.867} & \textbf{0.937} & \textbf{0.880} & \textbf{0.019} & \textbf{0.839} & \textbf{0.945} & \textbf{0.886} & \textbf{0.028} & \textbf{0.878} & \textbf{0.942} & \textbf{0.896}
                \\
                \bottomrule            
		\end{tabular}}  
\end{minipage}
\vspace{-4mm}
	\end{table*} 
\subsection{DeRUN} 
At stage $k$ and iteration $n$ ($1\leq n\leq N$), the gradient descent step of DeRUN dynamically adapts to the degradation severity, producing high‑quality results $\mathbf{X}_{k,n}$ without relying on any predefined degradation type. 
To generate the degradation prompt, we use a pretrained CLIP-based VLM, DA-CLIP~\cite{daclip} $DA\textbf{-}CLIP(\bigcdot)$, to infer degradation semantics, yielding a compact degradation-aware vector $\mathbf{d}_{k,n}$: 
\begin{equation}
    \mathbf{d}_{k,n} = {DA\textbf{-}CLIP}(\mathbf{X}_{k,n-1}),
\end{equation}
$\mathbf{d}_{k,n}$ guides the iterative restoration. 
We employ $\text{VLM}$-informed residual convolutional operators $RC_{k,n}^\mathbf{D}(\bigcdot)$ and ${RC_{k,n}^\mathbf{D}}^T(\bigcdot)$ (with a shared structure) to approximate the degradation operator $\mathbf{D}$ and its transpose $\mathbf{D}^T$~\cite{mou2022deep}:
\begin{equation}\label{eq:DeRUN2}
    RC_{k,n}^\mathbf{D}(\mathbf{X})=\bm{\sigma}_{k,n}\odot(CRC(\mathbf{X}) + \mathbf{X})+\bm{\mu}_{k,n},
\end{equation}
where $CRC(\bigcdot)$ denotes a Conv-ReLU-Conv block, and $\{\bm{\sigma}_{k,n}, \bm{\mu}_{k,n}\}$ are variational parameters:
\begin{equation}
    \bm{\sigma}_{k,n}=conv3(li(\mathbf{d}_{k,n})), \ \ \bm{\mu}_{k,n} = conv3(li(\mathbf{d}_{k,n})),
\end{equation}
with $li(\bigcdot)$ being a linear layer. \cref{eq:DeRUN2} seamlessly integrates the degradation prompt derived from the VLM into the iterative process. The gradient update at each step is given
\begin{equation}\hspace{-3mm}\label{eq:DeRUNGradient}
\begin{aligned}
    &\hat{\mathbf{X}}_{k,n}  \!=\! \hat{\mathcal{X}}(\mathbf{X}_{k,n-1}, \mathbf{X}_{k-1}),    \\
    & \!=\! \mathbf{X}_{k,n-1}\!-\!\alpha_{\mathbf{X}} {RC_{k,n}^\mathbf{D}}^T( RC_{k,n}^\mathbf{D}(\mathbf{X}_{k,n-1}) \!-\! \mathbf{X}_{k-1}),
\end{aligned}
\end{equation}
where $\mathbf{X}_{k-1}=\mathbf{X}_{k-1,N}$ for brevity. 
This operation can be interpreted as a progressive removal of residual degradation.

The proximal update is defined as follows:
\begin{equation}\label{eq:DeRUNProximal}
\begin{aligned}
    {\mathbf{X}}_{k,n} =& \mathcal{X}(\hat{\mathbf{X}}_{k,n}, {\mathbf{B}}_k,\mathbf{M}_k, \mathbf{Y}),\\
    =& \mathcal{X}_1(\hat{\mathbf{X}}_{k,n}) + \mathcal{X}_2({\mathbf{B}}_k, \mathbf{M}_k, \mathbf{Y}),    
\end{aligned}
\end{equation}
where $\mathcal{X}_1$ and $\mathcal{X}_2$ share the same structure as the lightweight network used in background estimation.
We introduce the complementary term $\mathcal{X}_2$, which leverages segmentation cues to highlight critical structural information and preserve discriminative details during enhancement. Empirically, uncertain and hard‑to‑segment regions usually correspond to ambiguous pixel‑level content; thus, this auxiliary term encourages DeRUN to focus on such challenging areas.

\subsection{BUI}
BUI exploits the nested structure by encouraging mutual guidance. SODUN masks provide structural priors to DeRUN (via $\mathcal{X}_2$ in~\cref{eq:DeRUNProximal}, while DeRUN's outputs are dynamically selected to guide subsequent segmentation.

Given the intermediate outputs $\{\mathbf{X}_{k-1,n}\}_{n=1}^N$, we employ a comprehensive Image Quality Assessment (IQA) scheme integrating TOPIQ~\cite{chen2024topiq}, Q-Align~\cite{wu2023q}, and MUSIQ~\cite{ke2021musiq} to evaluate perceptual quality:
\begin{equation}
    S(\mathbf{X}_{k-1,n})= IQA(\mathbf{X}_{k-1,n}).
\end{equation}

The highest quality image, $\mathbf{X}_{k-1}^{T1}$, is selected to replace $\mathbf{X}_{k-1}$ in all SODUN and DeRUN updates (See Alg. \ref{Alg:NUN}).

Although higher‑quality restored inputs generally enhance segmentation, a robust framework should remain stable under minor perturbations. Hence, we introduce a cross‑stage consistency (CSC) loss, $L_{csc}$, to enforce consistent predictions between the best and second‑best quality inputs, \textit{i.e.}, $\mathbf{M}_k$ (superscript $T1$ omitted) and $\mathbf{M}_k^{T2}$ at stage $k$. The formal definition of $L_{csc}$ is provided in \cref{sec:loss}.

\subsection{Loss function}\label{sec:loss}
The total loss function is formulated with a parameter $\epsilon$:
\begin{equation}
L_t = L_{basic} + \epsilon L_{csc},
\end{equation}
where $L_{basic}$ is the basic loss that jointly supervises segmentation accuracy and image restoration quality. The specific definitions of $L_{basic}$ and $L_{csc}$ are:
\begin{equation}\hspace{-3mm}
\scalebox{0.9}{$
\displaystyle
\begin{aligned}
		L_{basic}\!&=\!\sum_{k=1}^{K}\frac{1}{2^{K-k}}\!\left[L^w_{BCE}\!\left(\mathbf{M}_k,GT_s\right)\!+\!L^w_{IoU}\!\left(\mathbf{M}_k,GT_s\right)\right.\\
		& + \|{\mathbf{X}}_k-\mathbf{X} \|^2_2 ],
	\end{aligned}
    $}
\end{equation}
\begin{equation}\hspace{-3mm}
\scalebox{0.9}{$
\displaystyle
\begin{aligned}
L_{csc}\! =\! \sum_{k=1}^{K}\frac{1}{2^{K-k}}
\![L^w_{BCE}\!\left(\mathbf{M}_k,\mathbf{M}_k^{T2}\right)
+\!L^w_{IoU}\!\left(\mathbf{M}_k,\mathbf{M}_k^{T2}\right)],
	\end{aligned}
    $}
\end{equation}
where $L_{BCE}^w$ is the weighted binary cross-entropy loss and $L_{IoU}^w$ is the weighted intersection-over-union loss. $GT_s$ and $\mathbf{X}$ denote the ground truth of the mask and clean data. We use $\mathbf{X}_k$ instead of $\mathbf{X}_k^{T1}$ to implicitly encourage the network to generate higher‑quality outputs as the iterations progress.

Our loss enables NUN to learn degradation‑invariant representations while preserving fine‑grained structural fidelity and ensuring stable multi‑stage training.

\begin{table*}[tbp!]
\begin{minipage}{1\textwidth}
\setlength{\abovecaptionskip}{0cm} 
\centering
\caption{Efficiency comparison with existing cutting-edge methods on COD, where the size of the test image is $352\times 352$. } \label{table:efficiency}
\vspace{1mm}
\resizebox{\columnwidth}{!}{
\setlength{\tabcolsep}{1.6mm}
\begin{tabular}{l|ccc|ccc|ccc}
\toprule
\multirow{2}{*}{Efficiency} & \multicolumn{3}{c|}{ResNet50} & \multicolumn{3}{c|}{Res2Net50} & \multicolumn{3}{c}{PVT V2} \\ \cline{2-10}
                            & FocusDiff~\cite{zhao2025focusdiffuser} & RUN~\cite{he2025run}    & \cellcolor{c2!20}  NUN (Ours)  & FEDER~\cite{He2023Camouflaged}    & RUN~\cite{he2025run}      &\cellcolor{c2!20}  NUN (Ours)    & CamoFocus~\cite{khan2024camofocus}  & RUN~\cite{he2025run}   &\cellcolor{c2!20}  NUN (Ours)  \\ \midrule
Parameters (M) $\downarrow$             & 166.17   & 30.41  & \cellcolor{c2!20} 33.17 & 45.92    & 30.57     & \cellcolor{c2!20}  35.38 & 68.85     & 65.17 &   \cellcolor{c2!20} 65.67   \\
FLOPs (G) $\downarrow$                  & 7618.49  &    43.36  & \cellcolor{c2!20} 51.72   & 50.03    & 45.73    & \cellcolor{c2!20} 54.07    & 91.35       & 61.83 & \cellcolor{c2!20}  70.59  \\
FPS $\uparrow$                        & 0.23     & 22.75  & \cellcolor{c2!20}  20.10   & 14.02    & 20.26    & \cellcolor{c2!20} 18.08    & 9.63       & 15.82 & \cellcolor{c2!20} 13.86  \\ \bottomrule
\end{tabular}}
\end{minipage} \\ \vspace{1mm}
\begin{minipage}{0.327\textwidth}
\setlength{\abovecaptionskip}{0cm} 
		\setlength{\belowcaptionskip}{0cm}
	\centering
        \caption{ Results on clean TOD (\textit{GDD}). 
        } \label{table:TODQuanti}
	\resizebox{\columnwidth}{!}{
		\setlength{\tabcolsep}{1.7mm}
	\begin{tabular}{l|ccc}\toprule 

		 {Methods} & \cellcolor{gray!40}mIoU~$\uparrow$&\cellcolor{gray!40}$F_\beta^{max}$~$\uparrow$&\cellcolor{gray!40} $M$~$\downarrow$ \\ \midrule
        EBLNet~\cite{he2021enhanced} & 0.870                                 & 0.922                                 & 0.064   \\
        RFENet~\cite{fan2023rfenet} & 0.886 & 0.938 & 0.057  \\
	RUN~\cite{he2025run}  & {{0.895}} & {{0.952}} & {{0.051}}  \\ 
    \rowcolor{c2!20} NUN (Ours) & \textbf{0.903} & \textbf{0.954} & \textbf{0.049}\\
    \bottomrule  \end{tabular}}
\end{minipage}
\begin{minipage}{0.326\textwidth}
\setlength{\abovecaptionskip}{0cm} 
		\setlength{\belowcaptionskip}{0cm}
\centering
\caption{Results on clean PIS (\textit{ETIS}).
        } \label{table:PISQuanti}
	\resizebox{1\columnwidth}{!}{
		\setlength{\tabcolsep}{1.67mm}
	\begin{tabular}{l|ccc}
		\toprule 
		{Methods} 
		& \multicolumn{1}{c}{\cellcolor{gray!40}mDice~$\uparrow$} & \multicolumn{1}{c}{\cellcolor{gray!40}mIoU~$\uparrow$} & \multicolumn{1}{c}{\cellcolor{gray!40}$S_\alpha$~$\uparrow$} \\ \midrule
        PolypPVT~\cite{dong2023polyp}  & {{0.787}} & {{0.706}} & {{0.871}} \\
        MEGANet~\cite{bui2024meganet} & 0.739 & 0.665 & 0.836 \\
        RUN~\cite{he2025run}  & {{0.788}} &{{0.709}} & {{0.878}}   \\
        \rowcolor{c2!20} NUN (Ours) & \textbf{0.791} & \textbf{0.716} & \textbf{0.880} \\
	 \bottomrule                      
	\end{tabular}}
\end{minipage}
\begin{minipage}{0.353\textwidth}
\centering
	\setlength{\abovecaptionskip}{0cm} 
		\setlength{\belowcaptionskip}{0cm}
	\caption{Results on clean CDD (\textit{CDS2K}).  
    } \label{table:CDDQuanti}
	\resizebox{1\columnwidth}{!}{
		\setlength{\tabcolsep}{2mm}
		\begin{tabular}{l|cccccccc}
        \toprule 
Methods    & {\cellcolor{gray!40}$S_\alpha$~$\uparrow$} & {\cellcolor{gray!40}$M$~$\downarrow$} & {\cellcolor{gray!40}$E_\phi$~$\uparrow$} & {\cellcolor{gray!40}$F_\beta$~$\uparrow$} \\ \midrule
HitNet~\cite{hu2022high}     & 0.563 & 0.118 & 0.564  & {{0.298}}   \\
FEDER~\cite{He2023Camouflaged} & 0.538 & 0.070 & 0.586 &  0.288  \\
RUN~\cite{he2025run} & {{0.590}} &  {{0.068}} & {{0.595}}  & {{0.298}}  \\
\rowcolor{c2!20} NUN (Ours) & \textbf{0.595} & \textbf{0.066} & \textbf{0.598} & \textbf{0.302} \\
\bottomrule
\end{tabular}}
\end{minipage}
\vspace{-3mm}
	\end{table*} 
\begin{table*}[tbp!]
\begin{minipage}{\textwidth}
\setlength{\abovecaptionskip}{0cm} 
		\setlength{\belowcaptionskip}{-0.2cm}
		\centering
            \caption{Results on real-world camouflaged object detection with synthetic degradation data. 
            } \label{table:DegradCODQuanti}
            \vspace{1mm}
		\resizebox{\columnwidth}{!}{
			\setlength{\tabcolsep}{1.4mm}
			\begin{tabular}{l|c|cccc|cccc|cccc|cccc} 
				\toprule
				\multicolumn{1}{c|}{}                                        & \multicolumn{1}{c|}{}                           & \multicolumn{4}{c|}{\textit{CHAMELEON} }                                                                                                                                         & \multicolumn{4}{c|}{\textit{CAMO} }                                                                                                                                             & \multicolumn{4}{c|}{\textit{COD10K} }                                                                                                                                          & \multicolumn{4}{c}{\textit{NC4K} }                                                                                                                        \\ \cline{3-18} 
				\multicolumn{1}{l|}{\multirow{-2}{*}{Methods}} & \multicolumn{1}{c|}{\multirow{-2}{*}{Backbones}} & {\cellcolor{gray!40}$M$~$\downarrow$}                                  & {\cellcolor{gray!40}$F_\beta$~$\uparrow$}                               & {\cellcolor{gray!40}$E_\phi$~$\uparrow$}                               & \multicolumn{1}{c|}{\cellcolor{gray!40}$S_\alpha$~$\uparrow$}                                   & {\cellcolor{gray!40}$M$~$\downarrow$}                                  & {\cellcolor{gray!40}$F_\beta$~$\uparrow$}                               & {\cellcolor{gray!40}$E_\phi$~$\uparrow$}                               & \multicolumn{1}{c|}{\cellcolor{gray!40}$S_\alpha$~$\uparrow$}                                   & {\cellcolor{gray!40}$M$~$\downarrow$}                                  & {\cellcolor{gray!40}$F_\beta$~$\uparrow$}                               & {\cellcolor{gray!40}$E_\phi$~$\uparrow$}                               & \multicolumn{1}{c|}{\cellcolor{gray!40}$S_\alpha$~$\uparrow$}                                   & {\cellcolor{gray!40}$M$~$\downarrow$}                                  & {\cellcolor{gray!40}$F_\beta$~$\uparrow$}                               & {\cellcolor{gray!40}$E_\phi$~$\uparrow$}                               & \multicolumn{1}{c}{\cellcolor{gray!40}$S_\alpha$~$\uparrow$}                                   \\ \midrule 
                \multicolumn{18}{c}{Single Degradation: Low-Light}                                   \\ \midrule
                \multicolumn{1}{l|}{FEDER~\cite{He2023Camouflaged}} & \multicolumn{1}{c|}{ResNet50}  & 0.056 & 0.721 & 0.847 & 0.817 & 0.130 & 0.537 & 0.693 & 0.636 & 0.066          & 0.510          & 0.742          & 0.670          & 0.092 & 0.619 & 0.764 & 0.698 \\
                \multicolumn{1}{l|}{FSEL~\cite{sun2025frequency}} & \multicolumn{1}{c|}{ResNet50} & 0.134 & 0.221 & 0.424 & 0.485 & 0.175 & 0.217 & 0.417 & 0.462 & 0.092          & 0.213          & 0.518          & 0.513          & 0.146 & 0.253 & 0.463 & 0.492    \\ 
                \multicolumn{1}{l|}{RUN~\cite{he2025run}}  & \multicolumn{1}{c|}{ResNet50} & 0.057 & 0.729 & 0.841 & 0.813 & 0.124 & 0.579 & 0.737 & 0.657 & 0.063          & 0.539          & 0.785          & 0.684          & 0.089 & 0.643 & 0.795 & 0.713     \\
                \rowcolor{c2!20} NUN (Ours) & ResNet50 & \textbf{0.054} & \textbf{0.736} & \textbf{0.851} & \textbf{0.824} & \textbf{0.112} & \textbf{0.634} & \textbf{0.754} & \textbf{0.724} & \textbf{0.055} & \textbf{0.608} & \textbf{0.795} & \textbf{0.751} & \textbf{0.072} & \textbf{0.714} & \textbf{0.826} & \textbf{0.794}  
                \\ \midrule
                \multicolumn{18}{c}{Single Degradation: Hazy}                                   \\ \midrule
                \multicolumn{1}{l|}{FEDER~\cite{He2023Camouflaged}} & \multicolumn{1}{c|}{ResNet50}  & 0.074          & 0.641          & 0.744          & 0.753          & 0.134          & 0.543          & 0.677          & 0.656          & 0.060          & 0.525          & 0.722          & 0.728          & 0.082          & 0.675          & 0.746          & 0.754          \\
                \multicolumn{1}{l|}{FSEL~\cite{sun2025frequency}} & \multicolumn{1}{c|}{ResNet50} & 0.075          & 0.643          & 0.755          & 0.751          & 0.144          & 0.505          & 0,668          & 0.623          & 0.081          & 0.467          & 0.721          & 0.656          & 0.102          & 0.596          & 0.752          & 0.696          \\
                \multicolumn{1}{l|}{RUN~\cite{he2025run}}  & \multicolumn{1}{c|}{ResNet50} & 0.067          & 0.652          & 0.769          & 0.748          & 0.128          & 0.547          & 0.681          & 0.636          & 0.059          & 0.528          & 0.726          & 0.731          & 0.079          & 0.678          & 0.754          & 0.764          \\
                \rowcolor{c2!20} NUN (Ours) & ResNet50 & \textbf{0.059} & \textbf{0.706} & \textbf{0.815} & \textbf{0.805} & \textbf{0.120} & \textbf{0.574} & \textbf{0.704} & \textbf{0.685} & \textbf{0.053} & \textbf{0.598} & \textbf{0.780} & \textbf{0.743} & \textbf{0.072} & \textbf{0.699} & \textbf{0.811} & \textbf{0.782}
                \\ \midrule
                \multicolumn{18}{c}{Combined Degradation: Low-Resolution, Low-Light, and Hazy}                                   \\ \midrule
                \multicolumn{1}{l|}{FEDER~\cite{He2023Camouflaged}} & \multicolumn{1}{c|}{ResNet50}  & 0.129          & 0.422          & 0.669          & 0.563          & 0.169          & 0.391          & 0.591          & 0.529          & 0.104          & 0.322          & 0.657          & 0.539          & 0.147          & 0.408          & 0.645          & 0.554          \\
                \multicolumn{1}{l|}{FSEL~\cite{sun2025frequency}} & \multicolumn{1}{c|}{ResNet50} & 0.136          & 0.413          & 0.676          & 0.559          & 0.173          & 0.394          & 0.579          & 0.531          & 0.110          & 0.317          & 0.637          & 0.547          & 0.150          & 0.397          & 0.649          & 0.547          \\
                \multicolumn{1}{l|}{RUN~\cite{he2025run}}  & \multicolumn{1}{c|}{ResNet50} & 0.125          & 0.457          & 0.680          & 0.574          & 0.169          & 0.394          & 0.578          & 0.533          & 0.100          & 0.345          & 0.664          & 0.548          & 0.143          & 0.547          & 0.630          & 0.572          \\
                \rowcolor{c2!20} NUN (Ours) & ResNet50 & \textbf{0.071} & \textbf{0.629} & \textbf{0.750} & \textbf{0.729} & \textbf{0.137} & \textbf{0.463} & \textbf{0.619} & \textbf{0.599} & \textbf{0.068} & \textbf{0.448} & \textbf{0.686} & \textbf{0.643} & \textbf{0.096} & \textbf{0.572} & \textbf{0.716} & \textbf{0.681} \\  
                \arrayrulecolor{red!60!black}\specialrule{.08em}{.2ex}{.2ex}\arrayrulecolor{black}
                RUN~\cite{he2025run} & PVT V2 & 0.085 & 0.599 & 0.800 & 0.714 & 0.142 & 0.496 & 0.659 & 0.614 & 0.085 & 0.429 & 0.712 & 0.627 & 0.115 & 0.543 & 0.72 & 0.657  \\
                \rowcolor{c2!20} NUN (Ours) & PVT V2 &   \textbf{0.028} & \textbf{0.862} & \textbf{0.943} & \textbf{0.890} & \textbf{0.077} & \textbf{0.770} & \textbf{0.846} & \textbf{0.784} & \textbf{0.035} & \textbf{0.732} & \textbf{0.880} & \textbf{0.808} & \textbf{0.048} & \textbf{0.815} & \textbf{0.898} & \textbf{0.838}
                \\ \bottomrule            
		\end{tabular}}    
\end{minipage}
\vspace{-5mm}
	\end{table*}

\section{Experiment}
\noindent \textbf{Experimental setup}. Our NUN is implemented in PyTorch and trained on two NVIDIA RTX 4090 GPUs using Adam with momentum terms (0.9, 0.999). Following the common practice~\cite{fan2020camouflaged,he2025run}, we incorporate multi-level deep features extracted from encoder-shaped backbones into our framework. All input images are resized to $352\times 352$ during training and testing phases. The batch size is set to 36, and the initial learning rate is $1\times10^{-4}$, decreased by a factor of 0.1 every 80 epochs. The stage number $K$ of SODUN is fixed at 4. For the nested DeRUN, its stage number $N$ varies with $K$, \textit{i.e.}, $N\in\{4,3,3,2\}$ across the four stages.

\subsection{Comparative Evaluation}
Descriptions of the datasets and metrics are provided in the supplementary materials (Supp). For clean data, DeRUN, same as RUN~\cite{he2025run}, serves as a module that reconstructs the image given the estimated foreground and background.

\subsubsection{COS tasks with clean data}
\noindent \textbf{Camouflaged object detection}.
Following FEDER~\cite{He2023Camouflaged}, we assess the performance of our model on COD under three backbones: ResNet50 \cite{he2016deep}, Res2Net50 \cite{gao2019res2net}, and PVT V2 \cite{wang2022pvt}. As shown in \cref{table:CDDQuanti,table:efficiency}, our NUN shows superior robustness and generalization capability across all four datasets while maintaining computational efficiency. 

\noindent \textbf{Transparent object detection}. TOD plays a key role in autonomous driving and scene understanding. As shown in~\cref{table:TODQuanti}, our NUN surpasses existing methods across all datasets, highlighting its potential to advance vision-based perception systems for autonomous driving.

\noindent \textbf{Polyp image segmentation}. We evaluate our approach on polyp image segmentation using the \textit{CVC-ColonDB} and \textit{ETIS} datasets. Following RUN~\cite{he2025run}, PVT V2 is adopted as the default encoder. As shown in \cref{table:PISQuanti}, our NUN consistently achieves leading performance across all datasets. 

\begin{table*}[tbp!]
\begin{minipage}{0.43\textwidth}
	\setlength{\abovecaptionskip}{0cm} 
	\centering
        \caption{ Results on real-world degraded COD. 
        } \label{table:CODQuantiReal}
	\resizebox{\columnwidth}{!}{
		\setlength{\tabcolsep}{1.1mm}
	\begin{tabular}{l|ccc|ccc}\toprule 

		\multicolumn{1}{l|}{}                          & \multicolumn{3}{c|}{\textit{PCOD-LQ}} & \multicolumn{3}{c}{\textit{{MCOD-LQ}}}  \\ \cline{2-7} 
		\multicolumn{1}{l|}{\multirow{-2}{*}{Methods}} & \cellcolor{gray!40}mIoU~$\uparrow$&\cellcolor{gray!40}$F_\beta^{max}$~$\uparrow$&\cellcolor{gray!40} $M$~$\downarrow$& \cellcolor{gray!40}mIoU~$\uparrow$&\cellcolor{gray!40}$F_\beta^{max}$~$\uparrow$&\cellcolor{gray!40} $M$~$\downarrow$ \\ \midrule
        FEDER~\cite{He2023Camouflaged} & 0.104 & 0.585 & 0.481 & 0.159 & 0.440 & 0.438 \\
        FSEL~\cite{sun2025frequency} & 0.108 & 0.593 & 0.463 & 0.230 & 0.338 & 0.453 \\
        RUN~\cite{he2025run} & 0.102 & 0.653 & 0.499 & 0.045 & 0.490 & 0.480  \\
    \rowcolor{c2!20} NUN (Ours) & \textbf{0.061} & \textbf{0.674} & \textbf{0.657} & \textbf{0.034} & \textbf{0.556} & \textbf{0.522} \\
    \bottomrule  \end{tabular}}
\end{minipage}
\begin{minipage}{0.266\textwidth}
	\setlength{\abovecaptionskip}{0cm} 
	\centering
        \caption{ Results on degraded TOD. 
        } \label{table:TODQuantiSyn}
	\resizebox{\columnwidth}{!}{
		\setlength{\tabcolsep}{1.1mm}
	\begin{tabular}{l|ccc}\toprule 
		\multicolumn{1}{l|}{}                          & \multicolumn{3}{c}{\textit{GDD}}  \\ \cline{2-4} 
		\multicolumn{1}{l|}{\multirow{-2}{*}{Methods}} & \cellcolor{gray!40}mIoU~$\uparrow$&\cellcolor{gray!40}$F_\beta^{max}$~$\uparrow$&\cellcolor{gray!40} $M$~$\downarrow$ \\ \midrule
        EBLNet~\cite{he2021enhanced} & 0.786 & 0.873 & 0.108 \\
        RFENet~\cite{fan2023rfenet} & 0.835 & 0.896 & 0.081\\
        RUN~\cite{he2025run} & 0.862 & 0.920 & 0.070 \\
    \rowcolor{c2!20} NUN (Ours) & \textbf{0.877} & \textbf{0.928} & \textbf{0.062} \\
    \bottomrule  \end{tabular}}
\end{minipage}
\begin{minipage}{0.281\textwidth}
	\setlength{\abovecaptionskip}{0cm} 
	\centering
        \caption{ Results on degraded PIS. 
        } \label{table:PISQuantiSyn}
	\resizebox{\columnwidth}{!}{
		\setlength{\tabcolsep}{1.1mm}
	\begin{tabular}{l|ccc}\toprule 
		\multicolumn{1}{l|}{}                          & \multicolumn{3}{c}{\textit{ETIS}}  \\ \cline{2-4} 
		\multicolumn{1}{l|}{\multirow{-2}{*}{Methods}} & \cellcolor{gray!40}mDice~$\uparrow$&\cellcolor{gray!40}mIoU~$\uparrow$&\cellcolor{gray!40} $S_\alpha$~$\downarrow$ \\ \midrule
        PolypPVT~\cite{dong2023polyp} & 0.580 & 0.511 & 0.731 \\
        MEGANet~\cite{bui2024meganet} & 0.667 & 0.596 & 0.810\\
        RUN~\cite{he2025run} &  0.715 & 0.638 & 0.830\\
    \rowcolor{c2!20} NUN (Ours) & \textbf{0.722} & \textbf{0.647} & \textbf{0.836} \\
    \bottomrule  \end{tabular}}
\end{minipage}
\vspace{-3mm}
	\end{table*}
\begin{figure*}[t]
\setlength{\abovecaptionskip}{0cm}
	\centering
	\includegraphics[width=\linewidth]{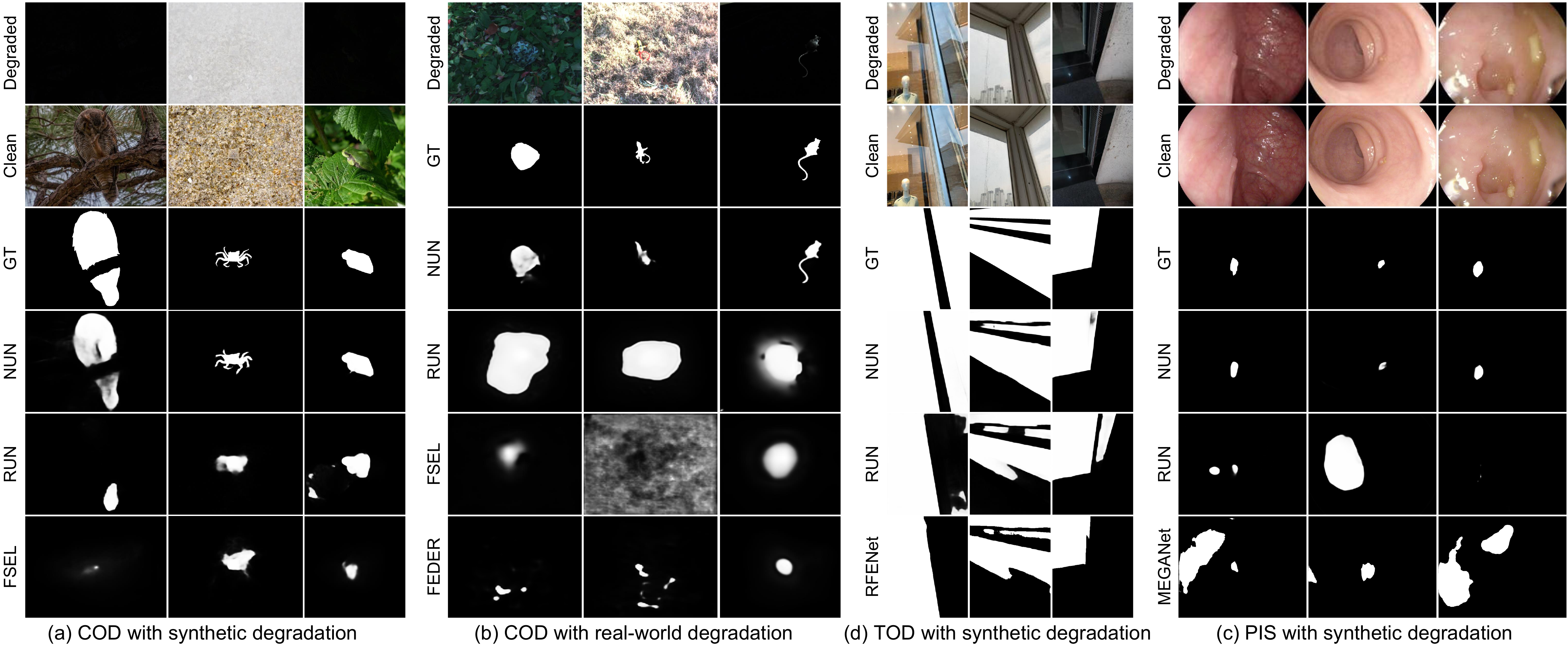}
	\caption{Visualization on COS tasks with real-world and synthetic degradation.
    }
	\label{fig:Visualization}
	\vspace{-3mm}
\end{figure*}
\begin{table*}[tbp!]
\begin{minipage}{0.445\textwidth}
	\setlength{\abovecaptionskip}{0cm} 
		\setlength{\belowcaptionskip}{-0.2cm}
	\centering
        \caption{ Breakdown ablations of NUN. 
        } \label{table:BreakdownAblation}
        	\resizebox{\columnwidth}{!}{
		\setlength{\tabcolsep}{1.1mm}
\begin{tabular}{cccc|cccc} \toprule
SODUN- & SODUN & DeRUN & BUI & $M$~$\downarrow$ & $F_\beta$~$\uparrow$ & $E_\phi$~$\uparrow$ & $S_\alpha$~$\uparrow$    \\ \midrule
$\checkmark$    & $\times$    & $\times$ & $\times$    & 0.102 & 0.328 & 0.603 & 0.583 \\
$\checkmark$    & $\checkmark$    & $\times$ & $\times$    & 0.093 & 0.362 & 0.619 & 0.592 \\
$\checkmark$    & $\checkmark$    & $\checkmark$ & $\times$    & 0.076 & 0.406 & 0.646 & 0.618 \\
\rowcolor{c2!20}$\checkmark$    & $\checkmark$    & $\checkmark$ & $\checkmark$  & \textbf{0.068} & \textbf{0.448} & \textbf{0.686} & \textbf{0.643} \\ \bottomrule
\end{tabular}}
\end{minipage}
\begin{minipage}{0.54\textwidth}
\setlength{\abovecaptionskip}{0cm} 
		\setlength{\belowcaptionskip}{-0.2cm}
	\centering
        \caption{ Ablation study of SODUN. 
        } \label{table:AblationSODUN}
        	\resizebox{\columnwidth}{!}{
		\setlength{\tabcolsep}{0.5mm}
\begin{tabular}{c|cccc|c} \toprule
{Metrics} 
& $\hat{\mathcal{M}}_1(\bigcdot) \!\rightarrow\! \hat{\mathcal{M}}(\bigcdot)$ & ${\mathcal{M}}_1(\bigcdot) \!\rightarrow\! {\mathcal{M}}(\bigcdot)$ & ${\mathcal{B}}_1(\bigcdot) \!\rightarrow\! {\mathcal{B}}(\bigcdot)$ & ${\mathcal{B}}_2(\bigcdot) \!\rightarrow\! {\mathcal{B}}(\bigcdot)$ &\cellcolor{c2!20} NUN (Ours) \\ \midrule
$M$~$\downarrow$ & 0.074 & 0.073 & \textbf{0.068} & 0.069 &\cellcolor{c2!20} \textbf{0.068}                  \\
$F_\beta$~$\uparrow$ & 0.412 & 0.416 & 0.447 & 0.445 &\cellcolor{c2!20} \textbf{0.448}                  \\
$E_\phi$~$\uparrow$ & 0.663 & 0.666 & \textbf{0.687} & 0.680 & \cellcolor{c2!20}0.686                  \\
$S_\alpha$~$\uparrow$  & 0.630 & 0.628 & 0.641 & 0.639 &\cellcolor{c2!20} \textbf{0.643}                  \\ \bottomrule 
\end{tabular}}
\end{minipage}\vspace{-5mm}
\end{table*}

\noindent \textbf{Concealed defect detection}. We assess the zero-shot generalizability to CDD, with models pre-trained on COD applied to \textit{CDS2K} to segment defects. Adhering to~\cite{he2025run}, PVT V2 serves as the default backbone. As verified by \cref{table:CDDQuanti}, our NUN achieves superior performance than others.

\subsubsection{COS tasks with degraded data}
\noindent \textbf{COD with synthetic degradation}. 
To simulate real-world conditions, we introduce three types of degradation (low-light, hazy, and low-resolution) across four datasets \cite{he2023degradation,he2023reti,he2025unfoldir} (see details in the Supp) and retrain all competing methods.
As shown in \cref{table:DegradCODQuanti,fig:Visualization}, FEDER \cite{He2023Camouflaged} and FSEL \cite{sun2025frequency} suffer performance degradation, while unfolding networks (RUN and our NUN) remain robust.
NUN generally outperforms RUN~\cite{he2025run} by $7.65\%$ (low-light), $6.90\%$ (hazy), and $20.42\%$ (combined degradation), highlighting strong resilience to real-world corruptions. 
With the stronger PVT‑V2 backbone, NUN achieves further gains, highlighting its practicality and potential for deployment.

\noindent \textbf{COD with real degradation}. To evaluate real-world generalization, we invite four annotators to build two low-quality subsets, \textit{PCOD-LQ} and \textit{MCOD-LQ}, from \textit{PCOD} \cite{wang2023polarization} and \textit{MCOD} \cite{li2025mcod}, yielding 160 high-consensus samples (\textit{PCOD-LQ}: 90; \textit{MCOD-LQ}: 70). See Supp for more details.
Models pre-trained on COD datasets with combined synthetic degradation are directly tested on the two datasets.
As shown in \cref{table:CODQuantiReal,fig:Visualization}, NUN achieves the best results, validating strong generalization to real degradations and the realism and effect of our synthetic degradation design.

\begin{table*}[tbp!]
\begin{minipage}{0.713\textwidth}
	\setlength{\abovecaptionskip}{0cm} 
		\setlength{\belowcaptionskip}{-0.2cm}
	\centering
        \caption{ Ablation study of DeRUN and BUI. 
        } \label{table:AblationDeRUN}
        	\resizebox{\columnwidth}{!}{
		\setlength{\tabcolsep}{1.1mm}
\begin{tabular}{c|ccc|ccc|c} \toprule
\multirow{2}{*}{Metrics} & \multicolumn{3}{c|}{DeRUN}  & \multicolumn{3}{c|}{BUI}                             &\cellcolor{c2!20} NUN \\ \cline{2-7}
&  $\hat{\mathcal{X}}_1(\bigcdot) \!\rightarrow\! \hat{\mathcal{X}}(\bigcdot)$ & ${\mathcal{X}}_3(\bigcdot) \!\rightarrow\! ({\mathcal{X}_1}(\bigcdot)\& {\mathcal{X}_2}(\bigcdot))$ & w/o ${\mathcal{X}_2}(\bigcdot)$ & w/o $IQA (\bigcdot)$ & $IQA_1(\bigcdot)\!\rightarrow\! IQA(\bigcdot)$ & w/o $L_{csc}$ &\cellcolor{c2!20} (Ours) \\ \midrule
$M$~$\downarrow$         & 0.073 & 0.068 & 0.071  &0.072 & 0.070 & 0.071 & \cellcolor{c2!20} \textbf{0.068}                 \\
$F_\beta$~$\uparrow$  &   0.422 & 0.452 & 0.430 & 0.421 & 0.435 & 0.428 &\cellcolor{c2!20} \textbf{0.448}                    \\
$E_\phi$~$\uparrow$  &    0.659 & 0.685 & 0.671 & 0.655 & 0.670 & 0.665 &\cellcolor{c2!20} \textbf{0.686}                  \\
$S_\alpha$~$\uparrow$   &   0.628 & 0.640 & 0.633 & 0.626 & 0.636 & 0.633 & \cellcolor{c2!20} \textbf{0.643}                  \\ \bottomrule    
\end{tabular}}
\end{minipage} 
\begin{minipage}{0.273\textwidth}
	\setlength{\abovecaptionskip}{0cm} 
		\setlength{\belowcaptionskip}{-0.2cm}
	\centering
        \caption{ Restored LQ data. 
        } \label{table:restoration}
        	\resizebox{\columnwidth}{!}{
		\setlength{\tabcolsep}{1.1mm}
\begin{tabular}{l|ccc} \toprule
Methods & PSNR~$\uparrow$   & FID~$\downarrow$                              & User Study \\ \midrule
Degraded &       11.17   &    95.63    &   1.17         \\
DiffIR~\cite{xia2023diffir} &  18.63                                &   81.07                               &    3.58        \\
RUN~\cite{he2025run}    & 15.21 & 90.12 &     2.25       \\
\rowcolor{c2!20}DeRUN  & 18.04 & 86.21 &     3.08       \\
\rowcolor{c2!20}NUN    & \textbf{22.37} & \textbf{73.15} &    \textbf{3.92}      \\ \bottomrule  
\end{tabular}}
\end{minipage} \\ \vspace{1mm}
\begin{minipage}{\textwidth}
	\setlength{\abovecaptionskip}{0cm} 
		\setlength{\belowcaptionskip}{0cm}
	\centering
        \caption{Integrating existing methods with different framework. ``-Res'': using a restoration method for pre-restoration. ``-BLO'' or ``-NUN'': integrating the method with bi-level optimization (jointly training with a restoration method)~\cite{he2023HQG} or our NUN.
        } \label{table:Integration}
        	\resizebox{\textwidth}{!}{
		\setlength{\tabcolsep}{1mm}
\begin{tabular}{c|cccc|cccc|cccc} \toprule
Metrics & FEDER & \multicolumn{1}{l}{FEDER-Res} & \multicolumn{1}{l}{FEDER-BLO} &\cellcolor{c2!20} FEDER-NUN & FSEL  & \multicolumn{1}{l}{FSEL-Res} & \multicolumn{1}{l}{FSEL-BLO} &\cellcolor{c2!20} FSEL-NUN & RUN   & \multicolumn{1}{l}{RUN-Res} & \multicolumn{1}{l}{RUN-BLO} &\cellcolor{c2!20} RUN-NUN  \\ \midrule
$M$~$\downarrow$ & 0.104 & 0.090                       & 0.082                       &\cellcolor{c2!20} \textbf{0.073}  & 0.110 & 0.093                      & 0.085                      &\cellcolor{c2!20} \textbf{0.078} & 0.100 & 0.088                     & 0.079                     &\cellcolor{c2!20} \textbf{0.070} \\
$F_\beta$~$\uparrow$& 0.322 & 0.357                       & 0.382                       &\cellcolor{c2!20} \textbf{0.416}  & 0.317 & 0.373                      & 0.412                      &\cellcolor{c2!20} \textbf{0.435} & 0.345 & 0.362                     & 0.391                     &\cellcolor{c2!20} \textbf{0.432} \\
$E_\phi$~$\uparrow$& 0.657 & 0.635                       & 0.660                       &\cellcolor{c2!20} \textbf{0.664}  & 0.637 & 0.652                      & 0.659                      &\cellcolor{c2!20} \textbf{0.668} & 0.664 & 0.657                     & 0.668                     &\cellcolor{c2!20} \textbf{0.673} \\
$S_\alpha$~$\uparrow$ & 0.539 & 0.586                       & 0.617                       &\cellcolor{c2!20} \textbf{0.635}  & 0.547 & 0.573                      & 0.607                      &\cellcolor{c2!20} \textbf{0.624} & 0.548 & 0.595                     & 0.618                     &\cellcolor{c2!20} \textbf{0.638} \\ \bottomrule
\end{tabular}}\vspace{-3mm}
\end{minipage}
\end{table*}

\begin{figure*}[t]
\begin{minipage}{0.49\textwidth}
\setlength{\abovecaptionskip}{0cm}
	\centering
	\includegraphics[width=\linewidth]{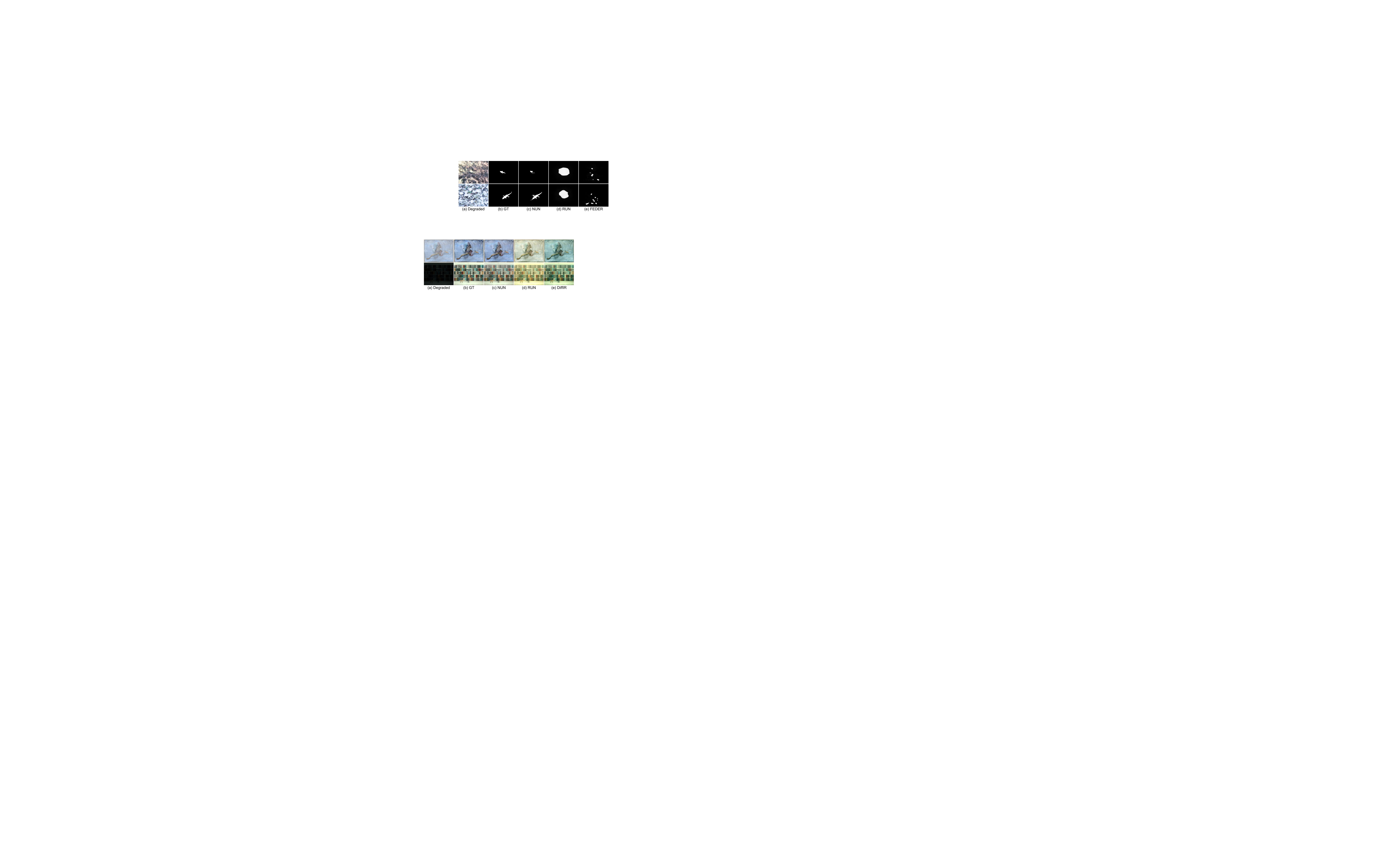}
	\caption{Segmentation performance in unseen degradation types.
 }
	\label{fig:UnseenDegradation}    \end{minipage} 
\begin{minipage}{0.49\textwidth}
\setlength{\abovecaptionskip}{0cm}
	\centering
	\includegraphics[width=\linewidth]{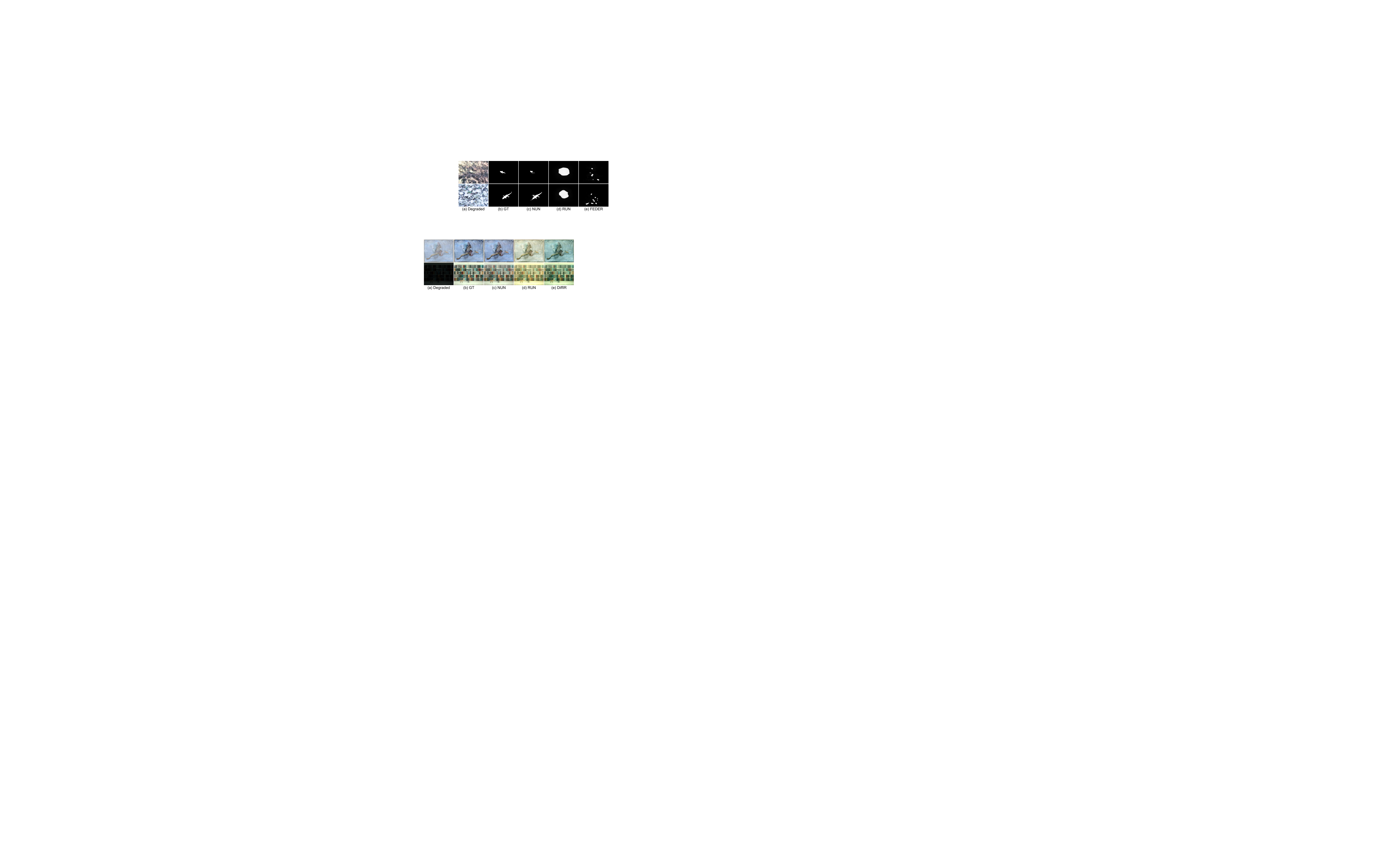}
	\caption{Image restoration of degraded data.
 }
	\label{fig:VisualAblation}
     \end{minipage}\vspace{-5mm}
\end{figure*}

\noindent \textbf{Other COS tasks with synthetic degradation}. 
We further extend our evaluation by introducing synthetic degradations into other COS tasks. 
We add different types of degradation to different tasks, accommodating their specific scenarios (see Supp.). All competing methods are retrained. As reported in~\cref{table:PISQuantiSyn,table:TODQuantiSyn,fig:Visualization}, our NUN achieves the best performance across all datasets, demonstrating its robustness and adaptability to diverse degradation scenarios.

\subsection{Ablation Study}\label{sec:ablation}
Unless otherwise specified, experiments of \cref{sec:ablation,sec:analysis} are conducted on \textit{COD10K} under combined synthetic degradations, including low‑resolution, low‑light, and hazy. Please see the ablations for the stage numbers in the Supp.

\noindent \textbf{Effect of SODUN}. As shown in~\cref{table:BreakdownAblation}, removing the background estimation part (SODUN-) leads to performance drop. Furthermore, in~\cref{table:AblationSODUN}, we analyze the impact of different architectural modifications: (1) adding extra regularization on~\cref{Eq:BasicModel1} like RUN~\cite{he2025run} ($\hat{\mathcal{M}}_1(\bigcdot) \!\rightarrow\! \hat{\mathcal{M}}(\bigcdot)$), changing the VSS module used in~\cref{eq:maskproximal} with a CNN module with similar parameters (${\mathcal{M}}_1(\bigcdot) \!\rightarrow\! {\mathcal{M}}(\bigcdot)$), replacing our simple network with a complex transformer~\cite{he2023reti} in~\cref{eq:backgroundproximal} (${\mathcal{B}}_1(\bigcdot) \!\rightarrow\! {\mathcal{B}}(\bigcdot)$), and removing $\mathbf{Y}$ in~\cref{eq:backgroundproximal} to break the clean-degradation pair. The performance decreases indicate the effect and efficiency of our SODUN.

\noindent \textbf{Effect of DeRUN}. The effect of DeRUN is jointly validated  in~\cref{table:BreakdownAblation,table:AblationDeRUN}. As shown in~\cref{table:AblationDeRUN}, several modifications lead to degraded performance: (1) replacing the explicit prior from DA-CLIP in~\cref{eq:DeRUNGradient} with implicit prior learned by the networks ($\hat{\mathcal{X}}_1(\bigcdot) \!\rightarrow\! \hat{\mathcal{X}}(\bigcdot)$); (2) replacing the light-weight network in~\cref{eq:DeRUNProximal} with more complex DiffIR used by RUN~\cite{he2025run} (${\mathcal{X}}_3(\bigcdot) \!\rightarrow\! ({\mathcal{X}_1}(\bigcdot)\& {\mathcal{X}_2}(\bigcdot))$); and (3) removing the segmentation cues in \cref{eq:DeRUNProximal} (w/o ${\mathcal{X}_2}(\bigcdot)$).
These consistent performance declines highlight the rationality and effectiveness of the proposed DeRUN design.

\noindent \textbf{Effect of BUI}. The BUI module is designed to enhance information exchange within the DUN‑in‑DUN architecture. As shown in \cref{table:BreakdownAblation,table:AblationDeRUN}, both components, the IQA‑based data selection strategy ($IQA_1(\bigcdot)$, which relies solely on MUSIQ for evaluation, similar to CORUN\cite{fang2024real}) and the consistency loss, contribute to performance improvement. They verify the superiority of our BUI module.

\subsection{Further Analysis and Applications}\label{sec:analysis}
Results on generalization to incomplete supervision and salient object detection are provided in the Supp. 

\noindent \textbf{Performance on unseen degradation types}. Our NUN trained on synthetic degradations generalizes well to real‑world scenarios with diverse and unseen degradation types.
As shown in \cref{table:CODQuantiReal,fig:UnseenDegradation}, NUN is robust to unseen degradation, like over‑exposure and snow.

\noindent \textbf{Image restoration quality}. 
As shown in \cref{table:restoration,fig:VisualAblation}, NUN achieves top results, with a user study containing 12 subjects.
In degraded COS, RUN \cite{he2025run} integrates DiffIR \cite{xia2023diffir} as its restoration backbone, yet its performance drops due to conflicts between restoration and segmentation goals.
In contrast, NUN alleviates such conflicts and surpasses both DiffIR and the standalone DeRUN, confirming its ability to jointly enhance restoration and segmentation.

\noindent \textbf{Integration with existing models}. 
NUN can be seamlessly integrated into conventional networks ( FEDER and FSEL) by replacing NUN's segmentation branch, and an unfolding‑based model (RUN) that embeds DeRUN within each unfolding stage. Results in \cref{table:Integration} reveal consistent performance gains, verifying our scalability and flexibility.

\noindent \textbf{Potential to serve as a unified vision system}. Existing strategies for real-world high-level vision tasks often rely on pre‑enhancement or joint optimization schemes such as bi‑level optimization (BLO)~\cite{he2023HQG} and the unfolding‑based integration (RUN).
While BLO promotes task‑specific optimization, and RUN attempts to balance segmentation and restoration objectives, both remain inferior to our unified formulation (\cref{table:Integration}).
Our experiments demonstrate that NUN not only attains superior segmentation and restoration results (\cref{table:restoration,fig:VisualAblation}) but also achieves mutual enhancement between the two.

\section{Conclusions}
We introduced NUN for real-world COS, which unifies image restoration and segmentation within a principled framework. By nesting DeRUN within SODUN and using VLM for degradation reasoning, NUN achieves strong robustness under diverse real-world degradations while maintaining efficiency. Experiments validate that NUN is a new SOTA in both clean and degraded scenes, revealing the potential of nested unfolding for broader visual perception tasks.

{
    \small
    \bibliographystyle{ieeenat_fullname}
    \bibliography{main}
}

\end{document}